\documentclass[%
 reprint,
 superscriptaddress,
nobibnotes,
 amsmath,amssymb,
 aip,
floatfix,
]{revtex4-2}

\usepackage{graphicx}
\usepackage{dcolumn}
\usepackage{bm}

\usepackage{physics}
\usepackage{hyperref}
\usepackage{xcolor}

\begin{document}


\title{SasAgent: Multi-Agent AI System for Small-Angle Scattering Data Analysis}

\author{Lijie Ding}
\email{dingl1@ornl.gov}
\affiliation{Neutron Scattering Division, Oak Ridge National Laboratory, Oak Ridge, TN 37831, USA}
\author{Changwoo Do}
\email{doc1@ornl.gov}
\affiliation{Neutron Scattering Division, Oak Ridge National Laboratory, Oak Ridge, TN 37831, USA}

\date{\today}

\begin{abstract}
We introduce SasAgent, a multi-agent AI system powered by large language models (LLMs) that automates small-angle scattering (SAS) data analysis by leveraging tools from the SasView software and enables user interaction via text input. SasAgent features a coordinator agent that interprets user prompts and delegates tasks to three specialized agents for scattering length density (SLD) calculation, synthetic data generation, and experimental data fitting. These agents utilize LLM-friendly tools to execute tasks efficiently. These tools, including the model data tool, Retrieval-Augmented Generation (RAG) documentation tool, bump fitting tool, and SLD calculator tool, are derived from the SasView Python library. A user-friendly Gradio-based interface enhances user accessibility. Through diverse examples, we demonstrate SasAgent’s ability to interpret complex prompts, calculate SLDs, generate accurate scattering data, and fit experimental datasets with high precision. This work showcases the potential of LLM-driven AI systems to streamline scientific workflows and enhance automation in SAS research.
\end{abstract}
\maketitle


\section{Introduction}
Small-angle scattering (SAS)\cite{lindner2024neutrons,windsor1988introduction}, including neutron\cite{chen1986small} and X-ray\cite{chu2001small} scattering, is an indispensable technique for probing nanoscale structures in materials such as polymers\cite{nierlich1979small}, proteins\cite{koch2003small,guilbaud2011using,mertens2010structural}, and nanoparticles\cite{li2016small,farrow2009relationship}, widely used in materials science\cite{gerold1978small,fratzl2003small,melnichenko2007small}, biophysics, and soft matter research\cite{wignall2005recent}. By measuring the scattering intensity $I(q)$ as a function of the scattering vector $q$, SAS provides critical insights into structural properties, serving as a powerful tool for material characterization and guiding the design of new materials. Despite its power, SAS data analysis presents significant challenges, requiring substantial expertise and time. Tasks such as selecting appropriate models from SasView’s library of 78 options\cite{archibald2020classifying,drucker2022challenges,tomaszewski2021machine,do2020small}, tuning model parameters, calculating scattering length density (SLD)\cite{windsor1988introduction} for samples and solvents, and fitting noisy experimental data demand manual intervention and specialized knowledge accumulated over years. These complexities, compounded by the need to interpret diverse datasets and avoid local minima in fitting, make SAS analysis labor-intensive and inaccessible to researchers lacking extensive experience.

Artificial intelligence (AI), particularly large language models (LLMs)\cite{radford2019language, xiao2025foundations}, offers a promising paradigm for scientific research by enabling intuitive automation of complex tasks like data analysis and tool integration. Multi-agent AI systems\cite{park2023generative, durante2024agent,xie2024large} comprising LLM-powered AI agent with specialized expertise work together like human teams, facilitate efficient workflows. The application of AI agent spans from software development\cite{yang2024swe}, chemical synthesis\cite{ramos2025review}, computer simulation\cite{mendible2025dynamate}, medical decision-making\cite{kim2024mdagents} and general research\cite{tang2025ai}. The application of LLM-based AI agents provides a promising path for automating and simplifying SAS data analysis. Meanwhile, adding the SAS analysis tool to the AI agent opens future possibility for agent-to-agent collaboration among AI agent with diverse expertise.

In this work, we address the gap between LLM agent and SAS data analysis, and address the aforementioned challenges of SAS data analysis by introducing SasAgent, a multi-agent AI system powered by LLMs, that leverages tools from the SasView software. SasAgent is self-aware, capable of performing three common SAS tasks: calculating SLD for specified samples, generating synthetic scattering data, and analyzing user-uploaded data based on text prompts. It can also guide users to interact with it effectively when prompted with general question. These capabilities are enabled by four LLM-powered AI agents, including a coordinator agent that directly interact with the user and three expert agents handling specialiszd tasks. The expert agents access four different tools built by wrapping SasView functions with LLM-friendly layers. We also build a web-based user interface using Gradio\cite{abid2019gradio}, enhancing user accessibility, allowing researchers to interact with SasAgent through intuitive text prompts and data uploads, and enabling hosting on personal web servers.

The rest of the paper is organized as the following: in Sec.~\ref{sec:system_design}, we describe the design of SasAgent in detail, and introduce the user interface. Four types of example use cases of SasAgent are covered in Sec.~\ref{sec:examples}. And finally, we summarize this work in Sec.~\ref{sec:summary}.

\section{System Design}
\label{sec:system_design}

\subsection{Multi Agent structure}
\begin{figure}[!h]
    \centering
    \includegraphics[width=\linewidth]{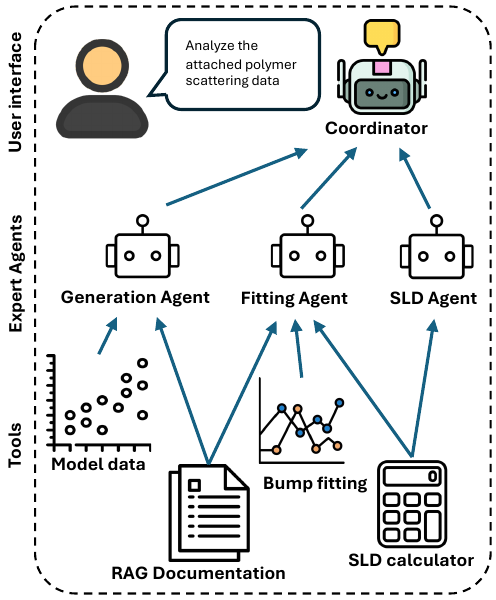}
    \caption{Architecture of the multi agent system that consists of three layers. In the user interface later, user will communicate with the coordinator agent through text, and user can also upload data. The coordinator will determine the type of task and distribute the task to the agents in the expert agents layer. Finally, the expert agents can use tools built from the SasView.}
    \label{fig:system_architecture}
\end{figure}

To enable our multi agent system to handle different kinds of request a user will have while minimizing the burden of single agent, we design a two layer structure among the agent as shown in Fig.~\ref{fig:system_architecture}. In total, we have 4 AI agents, all powered by LLM. The coordinator agent is responsible for direct communication with the user and distributing user requests to the appropriate expert agents. In the layer of expert agents, we create 3 different agents corresponding to 3 kinds of common tasks people carry out using SasView, and in the layer of tools, we build 4 different LLM-friendly tools for the agents to use. 

\begin{figure*}[t]
    \centering
    \includegraphics[width=\linewidth]{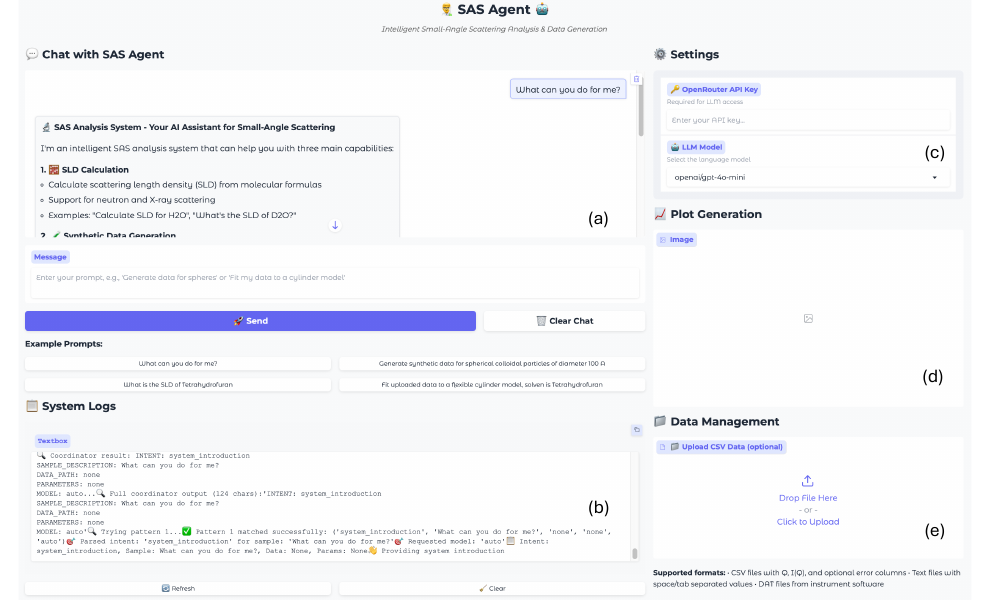}
    \caption{User interface of the SasAgent system consisting 5 blocks. (a) The chatting window in which the user can send prompt to the SasAgent, 4 example prompts are also displayed. (b) This is the system logs unit showing all the print out in the terminal, provide a under-the-hood view of our system. (c) is the setting block allow the users to put their API and choose preferred LLM. (d) The plotting window will show the generated plot. (e) The data manegement window is for the users to upload their data.}
    \label{fig:user_interface}
\end{figure*}

The 3 common tasks we consider are: scattering data generation, scattering data fitting, and SLD calculation. These tasks are handled by the generation agent, fitting agent, and SLD agent, respectively. In the generation task, the agent will process the user's request to figure out what kind of model to use for plotting the scattering intensity $I(q)$ versus scattering vector $q$ curve, determine the appropriate model parameter and q range based on the user prompt, and generate the scattering data and plot it. As for the fitting task, the fitting agent will read the user-uploaded scattering data, and determine the appropriate model for fitting this data based on user instruction, while use fixed SLD value for the solvent and sample and initial fitting guess for certain fitting parameters by interpreting user instruction. Finally, in the SLD calculation task, the SLD agent will calculate the real and imaginary part of the SLD for the user specified material.

To help the expert agent execute their tasks, we create 4 different tools by wrapping corresponding function within the SasView software. The model data tool allow the generation agent calculating the scattering function $I(q)$ curve using specified model and model parameters. In order to correctly use the model data tool, we build a Retrieval-Augmented Generation (RAG)\cite{lewis2020retrieval,gao2023retrieval} documentation tool that contains all of the documentation for all 78 models available in SasView, by crawling the documentation website\cite{sasview_fitting_models}. The model data tool and RAG documentation tool enable the generation agent to correctly identify variables and functions provided by SasView. We also have the bump fitting tool, which wraps the bump fitting function of the SasView for fitting execution. Finally, the SLD calculator tool is a wrapper of the SLD calculation function within SasView. The RAG documentation tool, bump fitting tool, and SLD calculator tool together power the fitting agent, as the agent read the documentation to understand how to use the model with bump fitting, and need to find the appropriate SLD for both sample and solvent as fixed parameter during the fitting procedure. The SLD calculator tool can also be used directly by the SLD agent to simply provide the SLD calculation function to the users.

In practice, our multi agent system is implemented using CrewAI\cite{crewai_framework}, and the RAG documentation is build with Beautiful soup\cite{richardson2007beautiful,nair2014getting}. Our LLM implementation uses OpenRouter\cite{openrouter_examples} to support multiple model choices.

\subsection{User interface}

To provide a user friendly interface for users to interact with the SasAgent system, we build a front end of our system using Gradio\cite{abid2019gradio}. Fig.~\ref{fig:user_interface} shows a snap shot of our user interface. The main part for interacting with our SasAgent is the chat window in Fig.~\ref{fig:user_interface}(a), where the user can just type in prompt and chat with our LLM-powered agent. On the bottom of the chat window, we add 4 example prompts for the user the try such as asking what the agent can do for the user, calculating SLD, generating synthetic scattering data, and analyzing uploaded scattering data. Below the chat window, in Fig.~\ref{fig:user_interface}(b), we have the system log section which displays the terminal printout text when running the system. This provides users with detailed insights into the agent system’s operations, including interactions between the coordinator and expert agents and how the expert agents utilize the tools, all shown in the system logs block.

Moving on to the right side of the user interface, we put the settings on the upper right as in Fig.~\ref{fig:user_interface}(c). The setting block allows users to add their own OpenRouter API if it is not configured in their environment or system setting. Since OpenRouter support almost every LLM available on the market, we add the model choice button allowing the user to choose different model, by default, we use the gpt-4o-mini model for it's low cost. Other available models included are: gpt-4o and gpt-5 from OpenAI\cite{achiam2023gpt}; claude-sonnet-4 from Anthropic\cite{anthropic2025claude4systemcard}; grok-3 and grok-4 from xAI; gemini-2.5-pro and gemini-2.5-flash from Google\cite{comanici2025gemini}. Bellow the LLM settings, we put the plot generation block for showing the figures generated by SasAgent in Fig.~\ref{fig:user_interface}(d) and the data management block for the user to upload files such as scattering data in Fig.~\ref{fig:user_interface}(e).

\section{Examples of system use case}
\label{sec:examples}
To demonstrate the usability of our SasAgent system, we provide few examples in the following. These examples include 4 main features such as the self-awareness of the system, the ability to calculate SLD from prompt, being able to generate synthetic scattering data using SasView models and fit user provided data.

\subsection{System guidance}
\begin{figure}[!h]
    \centering
    \includegraphics[width=\linewidth]{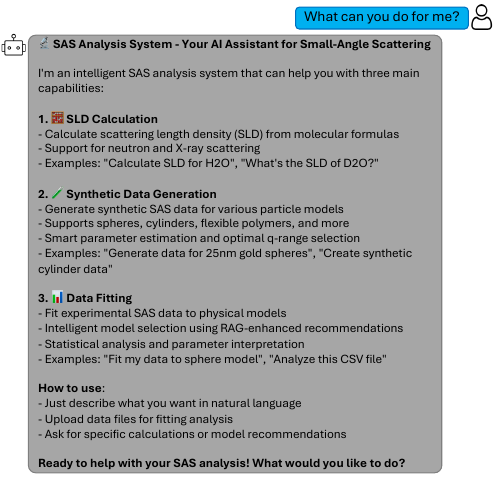}
    \caption{self-awareness of the SasAgent system, it provide guidance to the user by introducing itself and descibe the three main capabilities.}
    \label{fig:demo_introduction}
\end{figure}

While we have provided example prompt for the user to use the system, it is important to make the SasAgent self-aware, and being able to provide mode detailed guidance for users to use. As one of the example prompts, Fig.~\ref{fig:demo_introduction} shows that when asked about the capability of the SasAgent system, it replies by introducing itself and describe 3 main capabilities it can do for the user, while providing more detailed examples.

The system is capable of detecting user's intention from the input prompt, and will explain itself as it realizes a instruction is needed, such analysis is made possible by the LLM powering our system.

\subsection{Scattering length density calculation}
In small angle scattering experiment, it is often important to calculate the SLD of the sample or solvent before running the experiment or analyzing the experimental data. For neutron scattering, contrast matching is a crucial technique that purely depending on matching the SLD of the solvent with desired part of the sample. The estimation is usually done by using the corresponding tool in the SasView software. Here we demonstrate that by building such tool into the SasAgent system and allowing the LLM-powered agent to use this tool, we can simply find the SLD by prompting the SasAgent.

\begin{figure}[!h]
    \centering
    \includegraphics[width=\linewidth]{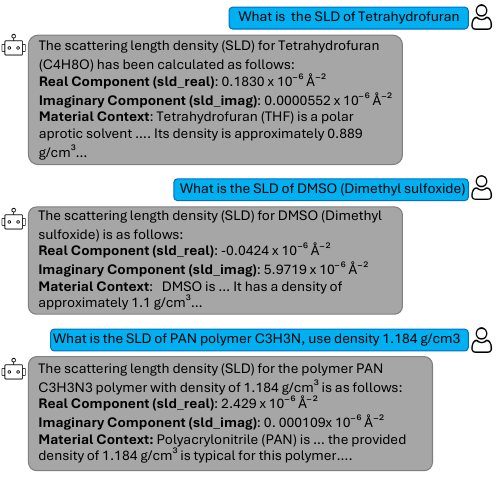}
    \caption{Calculating SLD calculation with the SasAgent. It analyze the user input and calls the sld calculator function wrapper for the SasView using appropriate input and output the corresponding results.}
    \label{fig:demo_sld_calculator}
\end{figure}

Fig.~\ref{fig:demo_sld_calculator} shows three examples of the SLD calculaton. In the first two examples, the user simply asks for the SLD by providing the name of the commonly known chemical such as Tetrahydrofuran and Dimethyl sulfoxide, and since the knowledge is already stored in the LLM, our agent was able to find the chemical formula and the density as the input for the SLD calculator tool and return both the real and imaginary part of the SLD to the user in the answer. In the third example, the use specify the chemical formula and density, and our SasAgent was able to directly extract these information to use as the input for the SLD calculator tool to find the SLD. In addition to the real and imaginary SLD, the agent also provides material context to the user, here in the example, the material context is shortened to only highlight that the agent finds and extracts the correct sample density for the SLD calculation.

This LLM-driven approach removes the manual step of looking up molecular data, lowering the barrier to entry for SLD estimation and accelerating workflows. As a result, the SLD agent not only supports the fitting agent in fixing solvent and sample SLD values during model fitting, but also serves as a user-friendly standalone assistant for rapid SLD queries in both neutron and X-ray scattering contexts

\subsection{Scattering data generation}
Another common use case is to visualize scattering data for specific model and related sample, such request is distributed to the generation agent, which has the model data tool to generate data and RAG documentation tool as a reference.

\begin{figure*}
    \centering
    \includegraphics[width=\linewidth]{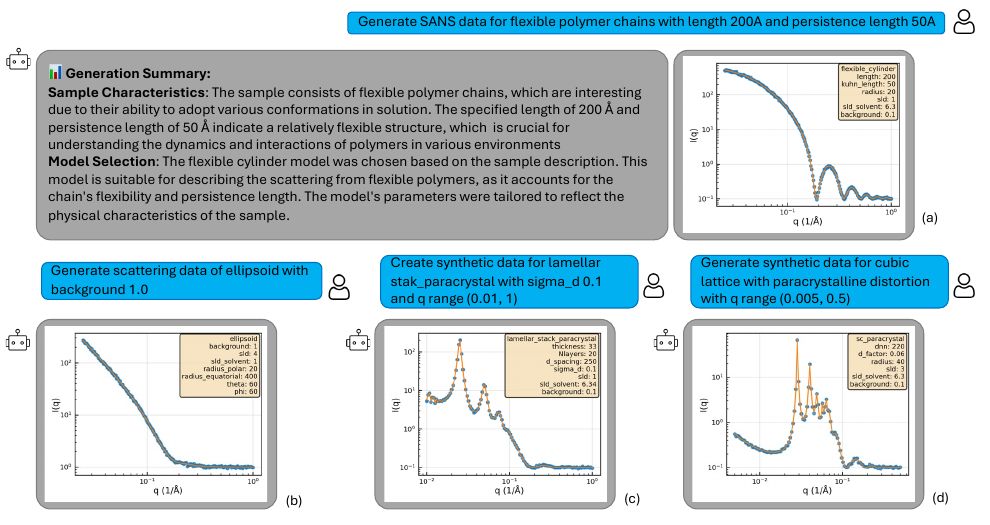}
    \caption{Generating scattering data with the SasAgent. (a) Asking the agent to generate the scattering of flexible polymer chain with specified length and persistence length. The agent chooses flexible cylinder model, plot the scattering curve and provide summary. (b) Example with ellipsoid model, only showing generated figure here for simplicity. (c) Example of a lamellar model with specified parameter and q range. (d) Example of a paracrystalline with specified q range.}
    \label{fig:demo_generation}
\end{figure*}

Fig.~\ref{fig:demo_generation} shows four examples of the scattering data generation usage of the SasAgent system. In Fig.~\ref{fig:demo_generation}(a), the user asks the SasAgent to generate small-angle neutron scattering data for the flexible polymer chains, and specify the length and persistence length. The agent generate the corresponding scattering data, plot and show it in the plot generation block, and give a summary about the data including the sample characteristics and model selection. Similarly Fig.~\ref{fig:demo_generation}(b)-(d) show another three example of the data generation, and for simplicity, we only show the final scattering plot here. In Fig.~\ref{fig:demo_generation}(b), the user ask for the scattering data of ellipsoid with specific background and no further detail, the system simply find the corresponding model and plot the scattering curve using default parameters in the model along with user-specified background level. In Fig.~\ref{fig:demo_generation}(c), the user asks for the scattering data for a lamellar model, and specified the $\sigma_d$ parameter and desired q range, the SasAgent system correctly understands the user's intention, set the $\sigma_d=0.1$ as specified by the user, and uses the correct q range $(0.01,1)$ according to user's description. Finally, in Fig.~\ref{fig:demo_generation}(d), the SasAgent system also correctly interpret the user's instruction, pick the correstion paracrystal model for the data generation, and use the correct q range based on user's description. While these four examples are not a exhausted list of all possible models and parameter can be interpreted by the SasAgent, they cover a diverse range of different material one may use with the SasAgent system.

\subsection{Scattering data fitting}
Lastly, a common, useful task can be done with the SasAgent system is the data fitting procedure. Under this scenario, the SasAgent will read the user uploaded file from the cache, and use the bump fitting tool to fit the user-provided scattering data. It is common practice that, when using the models in SasView to fit the scattering data, that researchers provide an estimation of the SLD for the sample and solvent and set these as fixed parameters to prevent the underlying regression model from getting stuck at unwanted local minima. Thus, in addition to the bump fitting tool to carry out the fitting and RAG documentation tool to understand how to use the specific models, we provide the SLD calculator tool to the fitting agent for determining or at least estimating the fixed SLD values. And for this task, we expect the user to provide basic information about the scattering data such as what the material of the sample, solvent, or expected range for certain parameters.

\begin{figure*}
    \centering
    \includegraphics[width=\linewidth]{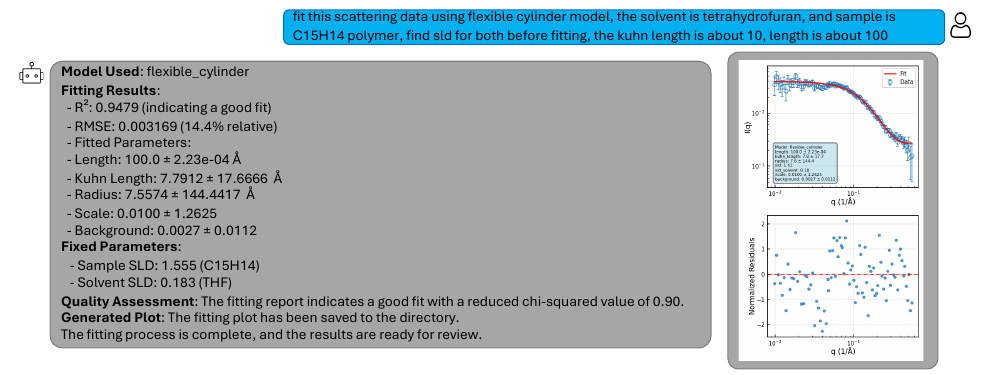}
    \caption{Fitting user provided scattering data with the SasAgent. With user provided specific model choice and the description of the solvent and sample, SasAgent choose the correct model and find the sld for the sample and the solvent and keep them fixed during the fitting procedure, and use user provided estimated parameters as initial guess, fit the provided scattering data using the bump fitting functionality of SasView.}
    \label{fig:demo_fitting_polymer}
\end{figure*}
Fig.~\ref{fig:demo_fitting_polymer} shows an example of fitting a scattering data of polymer chain using flexible cylinder model, the data is from the SANS experiments of a ladder polymer in solution sample\cite{ding2025ladder}. To help with the data fitting, in the user prompt, we tell the agent about the solvent and chemical formula of the sample, as well as estimation of the fitting parameter for the agent to use as initial guess. The agent system successfully identifies the user-uploaded data, recognizes it as a fitting task, and distributes the task to the fitting agent. The fitting agent firstly use the SLD calculator tool to find the fixed SLD values to use in the fitting procedure, then use the RAG documentation tool to identify the model along with its appropriate fitting parameters, finally the agent carry out the fitting procedure using bump fitting tool, using user specified initial values for the length and Kuhn length, and return the fitting results. The agent output contains two parts, one is the text summary contain the model used, fitting results, fixed parameters, and quality assessment such as the $\chi^2$ value of the fit, as shown on the left side of Fig.~\ref{fig:demo_fitting_polymer}. In addition, the agent return the plot of the fitted curve on top of user provided scattering data, as well as the normalized residuals. We note that for this specific ladder polymer sample, a sample-specific model is needed to fully capture the characteristics of scattering\cite{ding2025ladder}, this example is just to demonstrate the usage of SasAgent's fitting functionality.

\begin{figure}
    \centering
    \includegraphics[width=\linewidth]{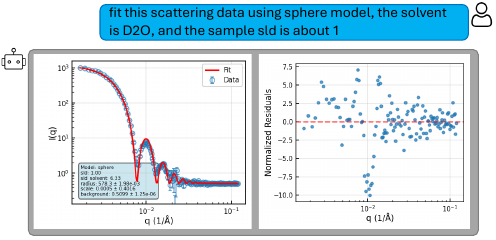}
    \caption{Fitting user provided scattering data for the colloids sample with the SasAgent. The system use user specified model and sample sld and solvent sld to fit the provided data.}
    \label{fig:demo_fitting_colloid}
\end{figure}
Fig.~\ref{fig:demo_fitting_colloid} shows another fitting result for a experimental data of dilute colloids solution sample\cite{tung2025colloids}. Since it is a dilute colloids, the user instruct the SasAgent to fit the data using sphere model, and inform the agent the solvent is heavy water, as the SLD for the sample is unknown, the user ask the agent to simply use 1 as an estimation. In practice, when the sample SLD is unknown, the scale of the fitting would not provide useful information such as the volume fraction. In this case, the agent successfully fit the scattering data using the sphere model, and find the radius about 57.83nm, or 115.66nm in diameter, which is consistent with the literature value 115nm\cite{tung2025colloids}.

These examples demonstrate that the SasAgent is capable of utilizing the tools migrated from the SasView to carry out data analysis works for small-angel scattering data, just like an scientist with good experience in the field.

\section{Summary}
\label{sec:summary}

In this work, we present a LLM-powered, multi-agent AI system that leverages tools from in the SasView software to carry out three common tasks relevant to small-angle scattering analysis. Our system is self-aware, and can execute tasks based on the prompts given by the user. These tasks include SLD calculation, synthetic data generation, and experimental data analysis. Each task corresponds to an expert agent that utilizes one or more tools to perform its tasks. These tools migrate from the functions available in the SasView, and modified to be LLM-friendly. A coordinator agent is in charge of all three expert agents, and directly communicate with the user. We build a web-based user interface to make our system easier to use, and we includes few examples for each use case. These examples demonstrate the robustness, diversity and flexibility of our SasAgent system.

As a first demonstration of integrating the LLM-based AI agent with the small-angle scattering tool, this work leaves space for future improvement. As our current implementation doesn't support the web search tool, its capability is limited by the LLM's knowledge. Future work should enable the web search tool so that the agent can better response to user using internet information. In addition, the current SasAgent only has documentation parsed from the SasView website, we can also add a research agent that allows agents to read user-uploaded literature to better assist users in a more relevant context.

Looking forward, LLM-based AI agent paves a potential new paradigm for scientific research. AI agents with scientific tools have huge potential in assisting scientists carrying out day-to-day research work. In principle, the user of the SasAgent can be another AI agent system utilizing different scientific tools. And it can become part of the autonomous lab\cite{szymanski2023autonomous,dai2024autonomous,schmidgall2025agent}, working with other AI agents.

\section*{Data Availability}
The code and data for this work are available at the GitHub repository \url{https://github.com/ljding94/SasAgent}

\section*{Author Contributions}
LD and CD conceived this work; LD designed, implemented and tested the multi agent AI system; LD and CD wrote and edited the manuscript.

\section*{Acknowledgment}
This research was performed at the Spallation Neutron Source, which is a DOE Office of Science User Facilities operated by Oak Ridge National Laboratory. This research was sponsored by the Laboratory Directed Research and Development Program of Oak Ridge National Laboratory, managed by UT-Battelle, LLC, for the US DOE.


\section*{References}
\bibliography{reference}

\onecolumngrid
\clearpage
\appendix

\end{document}